# Implementation of high-efficiency, lightweight residual spiking neural network processor based on field-programmable gate arrays[*]

HOU Yue[1], XIANG Shuiying[1,2], ZOU Tao[1], HUANG Zhiquan[1], SHI Shangxuan[1], GUO Xingxing[1], ZHANG Yahui[1], ZHENG Ling[3,1], HAO Yue[2]

1.State Key Laboratory of Integrated Service Networks, Xidian University, Xi'an 710071, China

2.State Key Discipline Laboratory of Wide Bandgap Semiconductor Technology, Xidian University, Xi'an 710071, China

3.School of Communication and Information Engineering, Xi'an University of Posts and Telecommunications, Xi'an 710121, China

**Abstract**

With the development of hardware-optimized deployment of spiking neural networks (SNNs), SNN processors based on field-programmable gate arrays (FPGAs) have become a research hotspot due to their efficiency and flexibility. However, existing methods rely on multi-timestep training and reconfigurable computing architectures, which increases computational and memory overhead, thus reducing deployment efficiency. This work presents an efficient and lightweight residual SNN accelerator that combines algorithm and hardware co-design to optimize inference energy efficiency. In terms of the algorithm, we employ single-timesteps training, integrate grouped convolutions, and fuse batch normalization (BN) layers, thus compressing the network to only 0.69M parameters. Quantization-aware training (QAT) further constrains all parameters to 8-bit precision. In terms of hardware, the reuse of intra-layer resources maximizes FPGA utilization, a full pipeline cross-layer architecture improves throughput, and on-chip block RAM (BRAM) stores network parameters and intermediate results to improve memory efficiency. The experimental results show that the proposed processor achieves a classification accuracy of 87.11% on the CIFAR-10 dataset, with an inference time of 3.98 ms per image and an energy efficiency of 183.5 FPS/W. Compared with mainstream graphics processing unit (GPU) platforms, it achieves more than double the energy

---

## 1. Introduction

With the rapid development of computer science and artificial intelligence (AI), artificial neural network (ANN) has been widely used in image processing, speech recognition and target tracking[1–5]. At the same time, with the continuous expansion of network scale, the demand for computing resources and energy consumption is increasing rapidly, and the high cost of computing and storage has become the main reason for limiting the development of ANN. In order to solve the problem of computing power and storage capacity, researchers have proposed spiking neural network (SNN), which is known as the third generation of neural network, by imitating the working mechanism of biological neural system[6]. Compared with ANN, the structure of SNN is closer to the human brain, and it uses pulse sequences to complete[7] information transmission between layers, which greatly improves the computational efficiency and energy efficiency optimization[8]. In addition, the sparsity of SNN makes neurons only active at specific time steps, and the network only computes when neurons are active, thus effectively reducing redundant computation, reducing hardware resource overhead, and significantly improving energy efficiency[9].

As a flexible and efficient hardware platform, field-programmable gate array (FPGA) has gradually become an ideal choice for SNN acceleration in recent years. Although the corresponding application specific integrated circuit (ASIC) may show superior performance and efficiency, such as IBM's TrueNorth[10] and Intel's Loihi[11]. However, each ASIC chip is designed according to specific requirements and tasks. Once the design is completed, the chip will be redesigned if it needs to be modified, which lacks flexibility and applicability[12]. On the other hand, FPGA can show higher flexibility and applicability[13] on the basis of ensuring efficient computing. Designers can redesign and write the hardware architecture according to the tasks and requirements to adapt to the rapid iterative network and algorithm updates and improve the performance of the accelerator[14]. Therefore, FPGAs are better suited than ASICs to handle SNN acceleration when flexible hardware architecture changes are required. In addition, due to the sparsity of SNN, most neurons remain static at most times, and only a small number of neurons emit pulse[9] at specific time points. FPGA can effectively utilize this sparsity, showing significant advantages in resource utilization and power consumption[15,16].

In recent years, based on the low-power, event-driven characteristics of SNN, many hardware architectures have been proposed[17–20] to achieve lower latency and higher energy efficiency. In 2022, Gerlinghoff et al. [21] proposed an end-to-end compilation framework E3NE supporting large-scale SNN. This framework adopts a new multi-level parallel coding scheme, which effectively improves computational efficiency, reduces hardware resource usage by over 50%, lowers power consumption by 20%, and shortens inference latency by an order of magnitude. In 2023, Chen et al.[22] proposed a high-performance general pulse convolution processing unit (SCPU) and its hardware architecture. The recognition rate of the processor on CIFAR-10 and CIFAR-100 data sets reached 92.45% and 68.55%, respectively, and the forward inference frame rate reached 40 frames/s, which significantly accelerated the inference process of deep SNN, especially on the residual block model. In 2024, Chen et al.[23] further proposed the SiBrain brain-like computing hardware architecture, whose core includes a sparse spatio-temporal parallel processing unit array for pulse convolution and pooling computing, and a fully connected core for pulse fully connected computing. The architecture achieves a classification accuracy of 90.25% on the CIFAR-10 dataset and an energy efficiency of up to 83GSOPs/W. In the same year, Aliyev et al. [24] first proposed a direct coding hybrid inference architecture, the main idea of which is to directly process the input layer data through a dense core and use a sparse core for event-driven pulse convolution calculation. This method has significant advantages over existing schemes in terms of throughput and power consumption.

Although these designs have made important progress in improving the performance of SNN hardware accelerators, there are still some limitations. For example, many schemes use multi-time-step training to improve the accuracy of the network, which increases the computational and storage overhead. In addition, most designs rely on reconfigurable computing architectures and require frequent access to off-chip memory, resulting in additional latency and power consumption. To address these issues, we implement a lightweight ResNet-10 spiking neural network and adopt a single-timestep training scheme to reduce computational complexity. Through strategies including group convolution, quantization-aware training (QAT), and batch normalization (BN) layer fusion, we compress the model parameters to 0.69 M. Furthermore, we implement the corresponding processor architecture on an FPGA platform. Within this architecture, the convolutional computing unit adopts the intra-layer resource multiplexing strategy to improve the resource utilization, combined with a full-pipeline architecture between layers to improve the computing throughput. Network parameters and calculation results are stored in on-chip block random access memory (BRAM). Finally, the proposed and implemented FPGA residual spiking neural network processor shows excellent performance in terms of inference time and energy efficiency.

## 2. Network Design and Lightweight Processing

SNN is inherently sparse and event-driven, which makes it computationally efficient on neuromorphic processors. However, SNN still faces many challenges in actual hardware deployment. On the one hand, the time dynamics significantly increase the computational complexity. On the other hand, the requirement to store membrane potential leads to a high memory footprint, which further restricts the energy efficiency performance of the system. In order to solve these problems, this paper introduces key technologies such as block convolution, time step reduction and quantization to deeply compress the network model, so as to effectively reduce the memory usage and computational overhead. These methods not only reduce data computation and storage requirements, but also improve inference efficiency, enabling deep SNNs to adapt to low-power hardware platforms more efficiently.

The network model used in this paper is a variant of ResNet-18. In order to reduce the scale of the model, the 64-channel and 512-channel parts of ResNet-18 are removed, and only 10 layers of trainable weights (excluding residual connection shortcuts) are retained, that is, the ResNet-10 spiking neural network. The network structure is shown in Fig. 1. Conv1 is a convolutional coding layer, which is responsible for converting the input into a pulse signal. Conv2 _ X and Conv3 _ X each contain two residual blocks, which are used as feature extraction modules and are mainly composed of a convolution layer and an activation layer; The pool layer is located before the fully connected (FC) layer, which compresses 256 feature maps into 256 × 1 vectors.

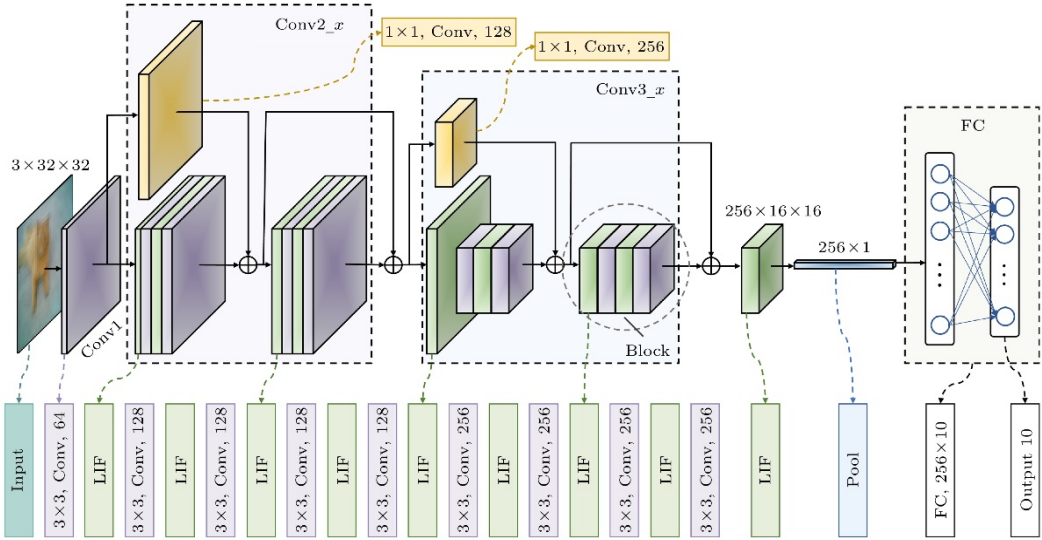

**Figure 1.** ResNet-10 spiking neural network structure.

The LIF (leaky integrate-and-fire) neuron model[25] is used as the active layer in the network. The LIF neuron model can be expressed as[26]:

$$\tau \frac{dU(t)}{dt} = -U(t) + RI_{in}(t), \qquad (1)$$

Where $I_{in}$ is the input stimulation current, $U$ is the membrane voltage, $R$ is the membrane resistance, and $\tau$ is the time constant of the circuit. When the weighted input from multiple presynaptic neurons exceeds the membrane potential, the neuron is excited and sends a pulse signal to the subsequent connection $S_{\text{out}}$:

$$S_{out}(t) = \begin{cases} 1, & \text{if } U(t) > U_{th}, \\ 0, & \text{otherwise}, \end{cases} \quad (2)$$

Where $U_{\text{th}}$ is the threshold of the membrane potential.

The LIF neuron model not only retains the biological characteristics such as leakage, integration and threshold release, but also greatly simplifies the computing model. Its operations mainly rely on addition and multiplication operations, which facilitates the deployment of large-scale neurons in hardware, especially in mobile devices and edge computing devices that require low resource consumption and low power consumption[27].

2.1 Grouped convolution strategy

Block convolution was first used to split the network on AlexNet to solve the problem of limited memory[28]. Now it is also widely used to reduce the number of parameters in the model, improve computational efficiency and save storage resources. In grouped convolution, the input and convolution kernels are grouped, and each convolution kernel only convolves with the corresponding feature map in the group. If the channel of the input feature map is $C_{\text{in}}$, the channel of the output feature map is $C_{\text{out}}$, the grouping parameter of the input data is $g$, and the size of the convolution kernel is $k \times k$, then the number of each group of input feature maps is $C_{\text{in}}/g$, and the number of output feature maps after convolution calculation is $C_{\text{out}}/g$. The traditional convolution parameter $P_c$ and the grouped convolution parameter $P_g$ can be calculated by (3) and (4) respectively to obtain[29]:

$$P_c = C_{in} \times C_{out} \times k \times k, \quad (3)$$

$$P_g = \frac{C_{in}}{g} \times \frac{C_{out}}{g} \times g \times k \times k. \quad (4)$$

The Fig. 2 compares the number of parameters in each layer when standard convolution and grouped convolution are used in the network respectively under the condition of one time step and $g = 4$, and the results show that the number of model parameters can be reduced to almost 1/4 of the original, only 0. 69 M.

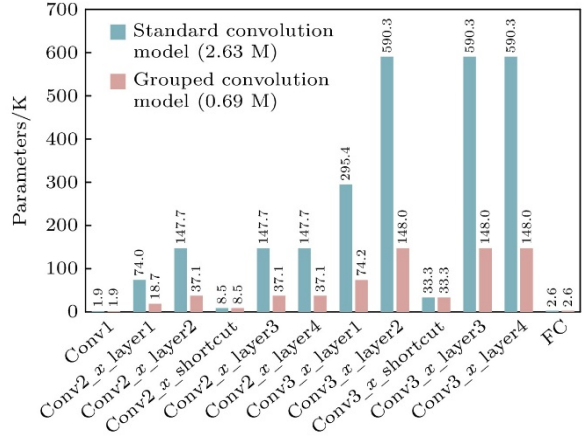

**Figure 2.** Comparison of parameter counts for each layer of ResNet-10 under standard convolution and group convolution.

2.2 Time Step Selection

As one of the important parameters of SNN, the size of time step $T$ directly affects the network performance. A larger $T$ value will increase the vertical depth of the network and enhance the information expression ability, but it will also significantly increase the computational overhead. A smaller $T$ value will reduce the complexity of the model and the number of parameters, but it will lead to a decline in accuracy. The test accuracy of the network under different conditions on the CIFAR-10 dataset[30] is compared Fig. 3. When the number of training rounds is 500, the maximum accuracy of the model with $T = 4$ is 91.40%, and the maximum accuracy of the model with $T = 1$ is 89.84%. Obviously, if only accuracy is considered, the $T = 4$ model performs better. However, when it comes to hardware deployment, fast inference speed and low power consumption are more critical. Although the accuracy of the $T = 1$ model is reduced, its activation process does not require complex calculation and membrane potential storage, but only the comparison of the weighted signal with the threshold. This not only significantly reduces hardware resource consumption and computing time, but also improves deployment efficiency. After comprehensive consideration, this paper finally chooses the model with $T = 1$.

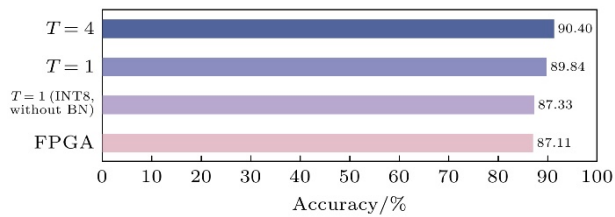

**Figure 3.** Comparison of test accuracy under different conditions.

2.3 BN layer fusion

In neural networks, the convolution layer is usually followed by a BN layer, which normalizes the input of each layer to concentrate the input value of the activation function in the sensitive interval of the nonlinear function, thus increasing the gradient and accelerating the convergence speed of the learning process. In order to further compress the network model, the BN layer is integrated into the convolution layer.

In the training process, the convolution operation can be expressed as:

$$y = w_i x_i + b, \tag{5}$$

Where $x_i (i = 1,2,\cdots,n)$ is the input signal of the $i$-th presynaptic neuro, $w_i (i = 1,2,\cdots,n)$ is the synaptic weight corresponding to the $i$-th input signal, $b$ represents the neuron bias, and $y$ is the output signal of the neuron.

The specific algorithm of BN layer operation is shown in (6) formula[31]. For a sample in a batch, assuming that the input of BN layer is $y$, the output after BN layer is

$$y_{bn} = \gamma \frac{y - \mu}{\sqrt{\sigma^2 + \varepsilon}} + \beta, \tag{6}$$

Where $\mu$ is the mean of the batch, $\sigma^2$ is the standard deviation, and $\varepsilon$ is a very small constant introduced to avoid divide-by-zero errors. The $\gamma$ and $\beta$ are variable parameters that are learned and updated during training just like the other parameters. However, in the inference process, these four data are fixed, and the BN layer can be fused into the convolution layer through linear calculation. The above BN layer calculation formula can be transformed into (7)[32]. In the inference process, the coefficients $c$ and $d$ are constants:

$$\begin{cases} y_{bn} = \gamma \dfrac{w}{\sqrt{\sigma^2 + \varepsilon}} x + (\beta + \gamma \dfrac{b - \mu}{\sqrt{\sigma^2 + \varepsilon}}) = cx + d, \\ c = \gamma \dfrac{w}{\sqrt{\sigma^2 + \varepsilon}}, \\ d = \gamma \dfrac{b - \mu}{\sqrt{\sigma^2 + \varepsilon}} + \beta. \end{cases} \tag{7}$$

Therefore, in network inference, the convolution weight can be adjusted from the original value to $c$, and the bias can be adjusted from $b$ to $d$, so as to realize the fusion of convolution layer and BN layer. Because the calculation process of BN layer is linear, this fusion method will hardly affect the accuracy of the model. In addition, after removing the BN layer, it is no longer necessary to perform calculations such as square root and division, which significantly simplifies the hardware design and reduces the computational and storage overhead, thereby reducing the overall energy consumption.

2.4 Parameter quantization

For quantization, the QAT[33] method is used to convert floating-point numbers in the network into fixed-point numbers, and pseudo-quantization nodes are inserted into the model to simulate the impact of quantization errors. Specifically, in the forward propagation process, the calculation uses the accuracy after quantization, while in the backward propagation, the gradient update still uses the floating point type to reduce the impact of quantization on the optimization process. The quantization operation follows (8)- (10)[34] in the forward propagation and (11) in the backward propagation[35]. Where $r$ is the floating-point number, $q$ is the converted fixed-point number, $S$ is the scaling factor, and $k$ is the bit width of the fixed-point number. Symmetric uniform quantization is used for parameter quantization, and the maximum absolute value needs to be calculated, that is, $r_{max}$.

$$q = round(S \times r), \tag{8}$$

$$r_{max} = \max(|r|), \tag{9}$$

$$S = 2^n, n = floor(\log_2(\frac{2^{k-1}}{r_{max}})), \tag{10}$$

$$\frac{\partial Loss}{\partial r} \stackrel{STE}{=} \frac{\partial Loss}{\partial q}. \tag{11}$$

Because of the discontinuity of the quantization operation, it is difficult to calculate its gradient directly, so the straight-through estimator (STE) is used to calculate the approximate gradient. By ignoring the effect of rounding on the gradient, STE allows the gradient to be passed directly to the unquantized variables, so that the model can still be trained by gradient descent[35]. This method effectively reduces the impact of quantization on model optimization, and improves the training stability and final accuracy of the quantized network.

In this implementation, all weights are quantized to 8 bits. It can be seen from the Fig. 3 that under the condition of single time step, the model test accuracy can reach 87.33% after 100 rounds of QAT fine-tuning with 8 bit accuracy by fusing BN layers. Compared with the single time step model without BN layer fusion and quantization, the accuracy is reduced by 2.51%. In practical applications, computational efficiency and energy efficiency are more important than absolute accuracy, so from the perspective of comprehensive performance, this paper selects the time step $T = 1$ in the algorithm, fuses the BN layer, and uses QAT to quantize the parameters to 8 bits, which improves the deployment efficiency and hardware friendliness while maintaining high accuracy.

**3. Design and Implementation of Processor Hardware System**

3.1 Processor overall framework design

The hardware architecture of the residual SNN processor designed in this paper is shown in Fig.

4. The core part of the architecture is the feature extraction module, including four Blocks, which operate in parallel to improve the computing speed and processing efficiency. Each Block contains two branch computation paths, one branch performs two activations and two pulse convolution operations, and the other branch performs residual convolution or direct mapping operations. The pooling module and the fully connected classification module are responsible for receiving and processing the pulse signal output from the feature extraction part, and finally generating the classification result. In the hardware design, all network parameters and intermediate calculation results are stored in the on-chip BRAM to ensure fast data access and avoid the access delay caused by the external memory. The controller is responsible for the scheduling of the whole system. The controller ensures that all modules work together through accurate input and output data scheduling.

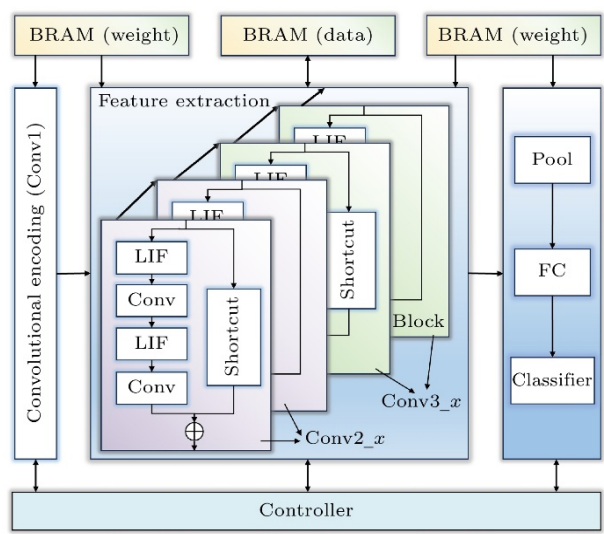

**Figure 4.** Overall hardware architecture of the residual SNN processor.

3.2 Design of Convolution Module in Core Part

In the network, the convolution operation occupies most of the computing and storage resources, so the optimization of convolution can significantly improve the processor efficiency. This paper uses the design of Chen et al.[36] in Eyeriss chip to improve the convolution module.

Fig. 5 shows a schematic of the convolution operation in the main path. Each processing element (PE) can perform a multiply-add operation, and nine PEs constitute a processing element core (PE core), which performs a 3 × 3 convolution kernel operation in one clock. The main processing element array (PE array) of the convolutional network consists of 64 PE cores. In the PE array, the weight and bias are loaded into the network from the BRAM in advance. After the weights are successfully loaded, the control signal initiates the reading of the feature map from BRAM. The data then undergoes padding and sliding-window processing before being fed into the PE array. There are 8 parallel output channels in the array, and each output channel can calculate the values of 8 input channels in parallel. For each output channel, the results from its corresponding 8 input channels are summed to produce the output feature map Fsum, which is then stored in the output BRAM. Because the output channels share the feature maps of the input channels, the feature maps flow into the PE Array and are broadcast to each output channel.

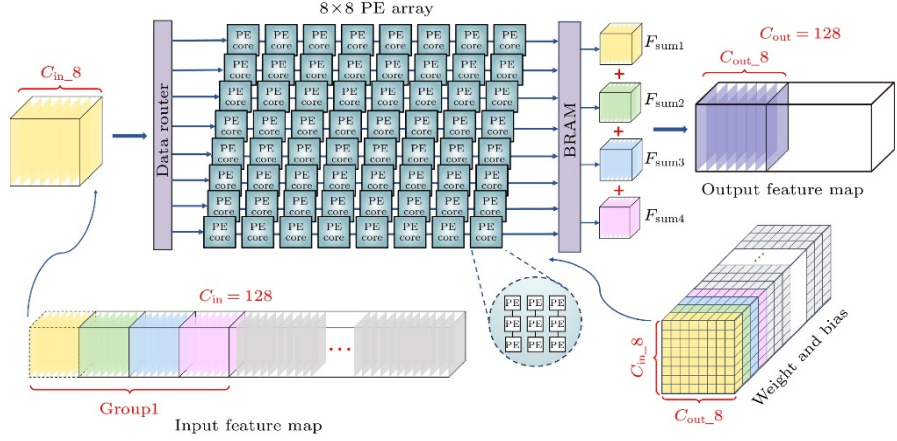

**Figure 5.** Schematic diagram of the convolution operation.

For the design of grouping convolution, taking the convolution of 128 channel input and 128 channel output as an example in the Fig. 5. The value of each output channel is derived from the convolution of the 32 input feature maps within the group with their corresponding kernel weights. Therefore, to compute every 8 output channels, the system reuses the PE array four times. The four intermediate results (Fsum1 to $Fsum_4$ in Fig. 5) are then summed to obtain the final result. In the convolution process, the controller is responsible for switching the input feature map and the convolution kernel parameters. The weight and the offset are stored in the BRAM according to the physical arrangement of the PE array, and the PE array is switched once when being multiplexed. Although the input feature map is also switched with the working beat of PE array, it is only used circularly within the group and will not be loaded from the external storage again before all the output feature maps of the same group are generated.

Because the network uses a fully pipelined architecture, the interlayer PE array structure is not universal. In the encoding layer, the size of the PE array is 3 × 64. Because the information transmitted is a non-pulse signal, the PE is composed of a digital signal processor (DSP) to perform multiply-add operations. In the main path convolution, the size of PE array is 8 × 8, and the information transmitted is a pulse signal without the participation of DSP. In the side residual convolution, the size of PE array is 64 × 8, and the PE is composed of DSP and performs multiply-add operation. Since the size of this part of convolution kernel is 1 × 1, each PE core contains only one PE.

3.3 Pipeline design

Data processing in convolution operation includes four independent steps: padding, input,

convolution computation and data output, which do not interfere with each other and have pipeline-like characteristics. Therefore, this paper uses pipelined optimized convolution computation. Fig. 6(a) is a schematic diagram of the pipeline structure in convolution operation data processing. It can be seen that the pipeline can reduce the delay and improve the system throughput without consuming too much resources. In addition, pipelining is applied not only to the convolution module but also to other components, resulting in a fully pipelined inter-layer architecture for the entire design. As shown in the Fig. 6(b), The full pipelined inter-layer architecture can improve the computing speed through inter-layer parallel computing. Under this architecture, the inference time of a single image is only 3.98 ms.

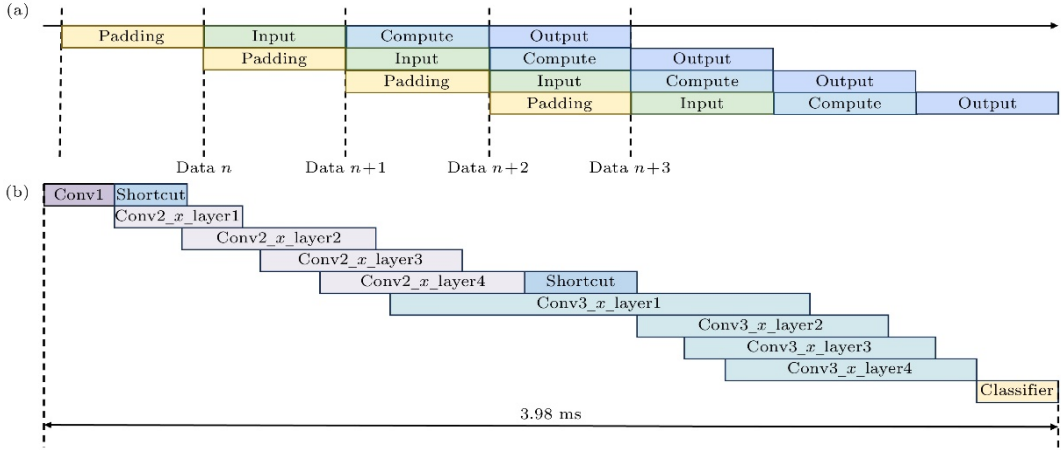

**Figure 6.** Pipeline design: (a) Convolution data processing pipeline structure; (b) fully pipelined inter-layer architecture.

## 4. Data analysis and discussion

The Tab. 1 shows the resource consumption of the processor on the FPGA, with DSP and BRAM accounting for the highest proportion. Because the processor uses a full on-chip memory strategy, the proportion of BRAM is large. At the same time, the amount of DSP resources directly affects the computing rate. In order to improve the inference speed, the processor also uses more DSP resources.

Table 1. Resource utilization of residual SNN processors

| Name | Consumed Resources | Available resources | Percentage/% |
|---|---|---|---|
| LUTs | 134859 | 425280 | 31.71 |
| FF | 341722 | 850560 | 40.18 |
| BRAM | 674.5 | 1080 | 62.45 |
| DSP | 3008 | 4272 | 70.41 |

The Tab. 2 is the result of comparing the processor and GPU (GeForce RTX 4060 Ti) on the CIFAR-10 data set[30]. Where FPS is the number of image frames per second that the processor can process, a measure of its computing speed. And FPS/W reflects the energy efficiency of the processor, which is the number of image frames that can be processed per 1 W of power consumed. It can be seen that the processor accuracy is reduced by 0.22% compared to the GPU. This gap is mainly due to the fact that only the weights and biases are quantized during training, while the input is also quantized during FPGA inference, resulting in a slight decrease in the final accuracy. On FPGA, its performance is evaluated by dynamic power consumption and single picture inference time. On the GPU, the inference time of a single image is calculated with the inference time of 100,000 images, and the real-time power consumption is measured to calculate its performance. The results show that FPGA can process a single image in 16 times the time of GPU, but the energy efficiency is improved by more than 2 times. It can be seen that even with limited resources, FPGAs still achieve high energy efficiency, demonstrate superior computing resource utilization, and are suitable for edge computing device deployment.

Table 2. Performance of the processor and GPU platform on the CIFAR-10 dataset

| Hardware platform | ZCU216 FPGA | GeForce RTX 4060 Ti |
|---|---|---|
| Accuracy/% | 88.11 | 88.33 |
| Power consumption/W | 1.369 | 51 |
| Single picture inference/ms | 3.98 | 0.243 |
| FPS | 251 | 4115 |
| FPS/W | 183.5 | 80.7 |

Tab. 3 compares the image recognition performance of the residual SNN processor designed and developed in this paper with other studies on the CIFAR-10 dataset. The data in the table are extracted from the original results of the references, or calculated and converted on the basis of the public data provided in the text to ensure fairness and comparability.

Table 3. Performance comparison with other SNN processors on the CIFAR-10 dataset

| Platform | E3NE[21] | SCPU[22] | SiBrain[23] | Aliyev et al.[24] | **This work** |
|---|---|---|---|---|---|
| FPGA model | XCVU13P | Virtex-7 | Virtex-7 | XCVU13P | **ZCU216** |
| Frequency/MHz | 150 | 200 | 200 | 100 | **100** |
| SNN model | AlexNet | ResNet-11 | CONVNet(VGG-11) | VGG-9 | **ResNet-10** |
| Model depth | 8 | 11 | 6(11) | 9 | **10** |
| Precision /bits | 6 | 8 | 8(8) | 4 | **8** |
| Number of parameters /M | - | - | 0.3(9.2) | - | **0.69** |
| LUTs/FFs | 48k/50k | 178k/127k | 167k/136k(140k/122k) | - | **135k/342k** |
| Accuracy /% | 80.6 | 90.60 | 82.93(90.25) | 86.6 | **87.11** |
| Power/W | 4.7 | 1.738 | 1.628(1.555) | 0.73 | **1.369** |
| Delay/ms | 70 | 25.4 | 1.4(18.9) | 59 | **3.98** |
| FPS | 14.3 | 39.43 | 696(53) | 16.95 | **251** |
| FPS/W | 3.0 | 22.65 | 438.8(34.1) | 23.21 | **183.5** |

E3NE[21] is an end-to-end FPGA acceleration framework for SNN inference. It combines efficient coding and hardware parallelization to achieve 99.1% accuracy on MNIST with an inference time of only 0.294 ms. However, when the network is deep, the convolutional layer and the pooling layer need to be multiplexed among multiple network layers, and the parameters are stored in off-chip memory, which increases the inference delay and reduces the energy efficiency. In the CIFAR-10 task, E3NE uses an 8-layer shallow network and 6-bit quantization. Although the resource overhead is small, the recognition accuracy is limited. By contrast, our processor employs a full-pipeline inter-layer architecture, 8-bit fixed-point quantization, and an improved ResNet-10 network. This combination achieves a 6.51% higher accuracy and delivers 61 times the energy efficiency of E3NE on the CIFAR-10 task. Although the use of LUT and FF is 2.8 times and 6.8 times that of E3NE, respectively, the overall energy efficiency advantage is still significant.

The SCPU[22] architecture consists of 256 parallel units, each of which integrates a LIF spiking neuron module and two computing paths, supporting standard pulse convolution and residual pulse convolution respectively, effectively improving the parallel computing capability. The design combines perceptual quantization and convolution fusion optimization, and skips membrane potential decay and threshold judgment to further speed up the calculation. However, this paper introduces the group convolution and single time step inference strategy to make the model more lightweight, This enables a 6× higher frame rate and an 8× improvement in energy efficiency while matching its accuracy.

SiBrain[23] architecture uses spatio-temporal parallel array, which supports multi-channel parallelism in space and can synchronously process four time steps in time dimension, thus significantly reducing inference latency. However, its weight storage relies on off-chip DDR, increasing access latency. SiBrain tested CONVNet and VGG-11 models on the CIFAR-10 data set. CONVNet is shallower and has better energy efficiency performance, but its accuracy is 4.18% lower than that of this paper. Under the VGG-11 network, although the accuracy of the processor in this paper is slightly lower, thanks to the single-time-step inference strategy and the on-chip storage mechanism, the image processing delay is effectively reduced, making the inference frame rate that is 4 times that of SiBrain, and an energy efficiency that is more than 5 times that of SiBrain.

The processor designed by Aliyev et al. [24] divides the network into sparse layer and dense layer, and flexibly allocates hardware resources according to the load demand. By compressing the pulse sequence (only recording the effective bit "1") and setting the clock gating (only providing the clock signal for the active area), quantizing the parameters to 4 bits, and using shift calculation instead of DSP, the power consumption is effectively reduced to 0.73 W. However, the architecture uses 4 bit accuracy and has a time step greater than 1, which is limited in accuracy and delay. In contrast, the processor in this paper achieves a 7.9-fold improvement in energy efficiency by reducing the parameter size and the amount of computation through group convolution while maintaining higher accuracy.

It can be seen that the processor in this paper significantly improves the operation efficiency by using the full-flow interlayer architecture, especially in terms of energy efficiency, on the premise of ensuring the same accuracy as other studies. In addition, the design in this paper is fully mapped on-chip, achieving excellent performance without off-chip memory access, which not only eliminates the additional power consumption caused by reconfigurable computing architecture in E3NE and other designs, but also greatly improves data throughput.

## 5. Conclusion

In this paper, an energy-efficient and lightweight residual SNN hardware accelerator is designed and implemented, which adopts the strategy of algorithm and hardware co-design to optimize the energy-efficient performance of SNN inference process. At the algorithm level, this paper implements a lightweight ResNet-10 spiking neural network suitable for hardware deployment, and uses a single time step to train the network. In this paper, the number of model parameters is reduced to 0.69 M by BN layer fusion, QAT quantization and group convolution. In terms of hardware design, this paper uses intra-layer resource reuse to improve FPGA resource utilization, adopts all-pipelined inter-layer architecture to improve computing efficiency, and leverages BRAM to store network parameters and computing results to reduce off-chip memory access. Finally, the processor is deployed on the ZCU216 FPGA platform. The experimental results show that the energy efficiency of the proposed processor is more than 2 times higher than that of the mainstream GPU platform. Furthermore, compared to other SNN processors, it delivers at least a 4× faster inference speed and a 5× higher energy efficiency. In the future, we can not only further explore the sparsity of weights and feature maps to compress the network scale, but also further enhance the full pipeline processing capacity of the network to shorten the inference delay, thus promoting the wide application of SNN hardware accelerators in low-power AI computing.

104202